%% file: ms.tex
\documentclass[letterpaper, 10 pt, conference]{ieeeconf}  %

\IEEEoverridecommandlockouts                              %

\input{preamble}

\usepackage[noadjust]{cite}

\title{\LARGE \bf
IntervenGen: Interventional Data Generation for \\
Robust and Data-Efficient Robot Imitation Learning
}

\author{Ryan Hoque$^{1,2}$, Ajay Mandlekar$^{*2}$, Caelan Garrett$^{*2}$, Ken Goldberg$^1$, Dieter Fox$^2$
\thanks{$^{1}$UC Berkeley, $^{2}$NVIDIA, *Equal contribution.}%
}

\begin{document}

\maketitle
\thispagestyle{empty}
\pagestyle{empty}

\begin{abstract}
    \input{chapters/abstract}
\end{abstract}
\input{chapters/intro}

\input{chapters/related_work}

\input{chapters/ps}

\input{chapters/method}

\input{chapters/setup}

\input{chapters/experiments}
\input{chapters/conclusion}

\input{acks.tex}

\bibliographystyle{IEEEtran}
\bibliography{IEEEabrv,references}

\clearpage

\end{document}

%% file: preamble.tex
\usepackage[tracking=true]{microtype}
\usepackage{amsmath} %
\usepackage{amssymb}  %
\usepackage{graphicx}
\usepackage{epsfig} %
\usepackage{mathptmx} %
\usepackage{times} %
\usepackage{amsmath} %
\usepackage{amssymb}  %
\usepackage{xspace}

\usepackage[frozencache,cachedir=minted]{minted}
\setminted{fontsize=\footnotesize}
\usepackage{url}
\usepackage[dvipsnames]{xcolor}
\usepackage[font={footnotesize}]{caption}
\usepackage{booktabs} %
\usepackage{tikz}
\usetikzlibrary{positioning,shapes.geometric,arrows,fit,shapes.symbols,svg.path,tikzmark,calc}
\usepackage{booktabs}
\usepackage{textcomp}
\usepackage{listings}
\usepackage{subcaption}
\pgfdeclarelayer{background}
\pgfdeclarelayer{foreground}
\pgfsetlayers{background,main,foreground}
\usepackage{siunitx}
\usepackage{adjustbox}
\usepackage{array}
\usepackage{multirow}
\usepackage{geometry}

\let\labelindent\relax %

\usepackage[inline]{enumitem} %

\usepackage{xspace}
\newcommand\algname{{IntervenGen}\xspace}
\newcommand\algabbr{{I-Gen}\xspace}

\newcommand{\revision}[1]{{#1}}
\newcommand{\ryan}[1]{}
\newcommand{\caelan}[1]{}
\newcommand{\ajay}[1]{}

\usepackage{algorithm}
\usepackage[noend]{algpseudocode}
\usepackage{amssymb}
\algnewcommand\algorithmicdeclare{\textbf{Declare:}}
\algnewcommand\Declare{\item[\algorithmicdeclare]}

\usepackage{booktabs}
\usepackage{hyperref}
\usepackage{dblfloatfix}
\usepackage{graphicx} 
\usepackage{caption}
\usepackage{pifont}
\usepackage{verbatimbox}
\usepackage{bm}

\newcommand{\proc}[1]{\textsc{#1}}

%% file: chapters/abstract.tex
Imitation learning is a promising paradigm for training robot control policies, but these policies can suffer from distribution shift, where the conditions at evaluation time differ from those in the training data. %
A popular approach for increasing policy robustness to distribution shift is interactive imitation learning (i.e., DAgger and variants), where a human operator provides corrective interventions during policy rollouts. However, collecting a sufficient amount of interventions to cover the distribution of policy mistakes can be burdensome for human operators. We propose \algname (\algabbr), a novel data generation system that can autonomously produce a large set of corrective interventions with rich coverage of the state space from a small number of human interventions. We apply \algabbr to 4 simulated environments and 1 physical environment with object pose estimation error and show that it can increase policy robustness by up to 39$\times$ with only 10 human interventions. Videos and more results are available at \url{https://sites.google.com/view/intervengen2024}.

%% file: chapters/intro.tex
\section{Introduction}\label{sec:intro}

Imitation Learning (IL) from human demonstrations is a promising paradigm for training robot policies. One approach is to collect a set of offline task demonstrations via human teleoperation~\cite{mandlekar2018roboturk, mandlekar2019scaling} and employ behavior cloning (BC)~\cite{pomerleau} to train robot policies via supervised learning, where the labels are robot actions. 
There have been recent efforts to scale this approach by collecting thousands of demonstrations using hundreds of human operator hours and training high-capacity neural networks on the large-scale data~\cite{Jang2022BCZZT, Brohan2022RT1RT, ebert2021bridge, ahn2022can, Lynch2022InteractiveLT}.

However, IL policies can suffer from distribution shift, where the conditions at evaluation time differ from those in the training data~\cite{ross2011reduction}. As an example, consider a policy that makes decisions based on object pose observations.
A common source of distribution shift in the real world is object pose estimation error, which can occur due to a wide range of factors such as sensor noise, occlusion, network delay, and model misspecification.
This can cause inaccuracy in the robot's belief of where critical objects are located in the environment, leading the robot to visit states outside the training distribution that result in poor policy performance.

One approach to addressing distribution shift is to collect a large set of demonstrations under diverse conditions and hope that agents trained on this data can generalize. However, human teleoperation data is notoriously difficult to collect due to the human time, effort, and financial cost required~\cite{Jang2022BCZZT, Brohan2022RT1RT, ebert2021bridge, ahn2022can, Lynch2022InteractiveLT}.

\input{figures_tex/pull}
An alternative approach is interactive IL (i.e., DAgger~\cite{ross2011reduction} and variants~\cite{Kelly2018HGDAggerII, Mandlekar2020HumanintheLoopIL, hoque2021thriftydagger}), where humans can intervene during robot execution and demonstrate \textit{recovery behaviors} to help the robot return to the support of the training distribution. Subsequent training on these corrections can increase policy robustness and performance both theoretically and in practice \cite{ross2011reduction}. However, interactive IL imposes even more burden on the human supervisors than behavior cloning, as the human must continuously monitor robot task execution and intervene when they see fit, typically over multiple rounds of interleaved data collection and policy training. Moreover, a significant amount of recovery data may be required to adequately cover the distribution of mistakes the policy may make.

We raise the following question: do we actually need to have a human operator collect corrections every single time a policy makes a mistake?
MimicGen~\cite{mimicgen}, a recently proposed data generation system, raises an intriguing possibility: a large dataset of synthetically generated demonstrations derived from a small set of human demonstrations (typically 100$\times$ smaller or more) can produce performant robot policies. 
The system's key insight is that similar object-centric manipulation behaviors can be applied in new contexts by appropriately transforming demonstrated behavior to the new object frame. 
\revision{Inspired by this insight,} we propose a data generation system for \textit{interventional} data (see Fig.~\ref{fig:teaser}). With a small set of corrective interventions from a human operator, we can autonomously generate data with significantly higher coverage of the distribution of potential policy mistakes. 
\revision{Our system can be applied to a broad range of applications such as improving policy success rates on a task of interest, making policies robust to errors in perception, and more broadly, acting as a domain randomization~\cite{tobin2017domain} procedure to aid in sim-to-real transfer of IL policies without requiring additional data collection from a human supervisor. In this work, we focus on improving policy robustness to errors in perception.}

\textbf{We make the following contributions:}

\begin{enumerate}
    \item \algname (\algabbr), a system for automatically generating interventional data across diverse scene configurations and broad mistake distributions from a small number of human interventions. 
    \item An application of \algabbr to improve policy robustness against 2 sources of object pose estimation error (sensor noise and geometry error) in 5 high-precision 6-DOF manipulation tasks. \algabbr increases policy robustness by up to 39$\times$ with only 10 human interventions.
    \item Experiments demonstrating the utility of \algabbr over alternate uses of a human data budget of equivalent or even greater size. A policy trained on synthetic \algabbr data from 10 source human interventions can outperform one trained on even 100 human interventions by 24\%, with 12\% of the data collection time and effort.
    \item \revision{An experiment that shows that policies trained in simulation with \algabbr are amenable to real-world deployment and retain robustness to erroneous state estimation}.
\end{enumerate}

%% file: figures_tex/pull.tex
\begin{figure}[ht!]
  \begin{center}
    \includegraphics[width=\linewidth]{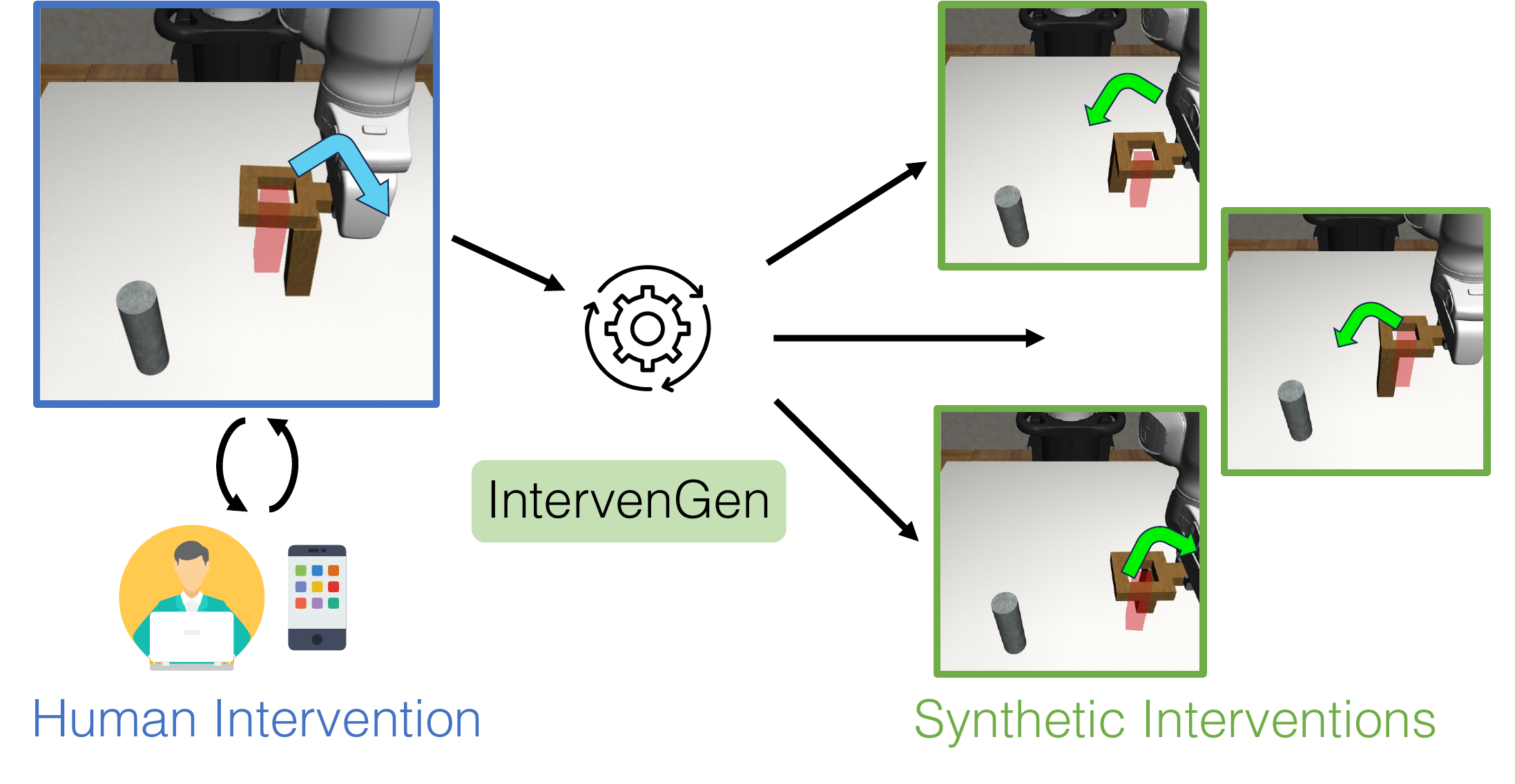}
  \end{center}
  \caption{\textbf{Overview.} \algname automatically generates corrective interventional data from a small number of human interventions, with coverage across both diverse scene configurations and policy mistake distributions. Here, the robot mistakenly believes the peg is at the position highlighted in red and requires demonstration of recovery behavior toward the true peg position.}
  \label{fig:teaser}
  \vspace{-15pt}
\end{figure}

%% file: chapters/related_work.tex
\section{Related Work}

\input{figures_tex/example}

\textbf{Data Collection Approaches for Robot Learning.}
Many prior works address the need for large-scale data in robotics. 
Some use self-supervised data collection~\cite{kalashnikov2021mt, dasari2019robonet}, %
but the data can have low signal-to-noise ratio due to the trial-and-error process. 
Other works collect large datasets using experts that operate on privileged information available in simulation~\cite{jiang2022vima, dalal2023imitating, mcdonald2022guided}. %
Still, designing such experts can require significant engineering. One popular approach is to collect demonstrations by having human operators teleoperate robot arms~\cite{mandlekar2018roboturk, mandlekar2019scaling, Brohan2022RT1RT, Jang2022BCZZT}; %
however, this can require hundreds of hours of human operator time. Some systems also allow for collecting interventions to help correct policy mistakes~\cite{luo2021robust, Mandlekar2020HumanintheLoopIL, liu2022robot}. In this work, we make effective use of a handful of interventional corrections provided by a single human operator to autonomously generate large-scale interventional data, substantially reducing the operator burden.

\textbf{Imitation Learning from Human Demonstrations.}
Behavioral Cloning (BC)~\cite{pomerleau} on demonstrations collected using robot teleoperation with human operators has shown remarkable performance in solving real-world robot manipulation tasks~\cite{zhang2017deep, mandlekar2020learning, Mandlekar2021WhatMI, Brohan2022RT1RT, Jang2022BCZZT, ahn2022can}. However, scaling this paradigm can be costly due to the need for large amounts of data, requiring many hours of human operator time~\cite{Brohan2022RT1RT, Jang2022BCZZT, Lynch2022InteractiveLT}. Furthermore, policies trained via IL are often brittle and can fail when deployment conditions change from the training data~\cite{ross2011reduction}. 

\textbf{Interactive Imitation Learning.}
Interactive IL allows demonstrators to provide corrective supervision in situations where policies require assistance. 
Some approaches require an expert to relabel states encountered by the agent with actions that the expert would have taken~\cite{ross2011reduction, chernova2009interactive}, %
but it can be difficult for human supervisors to relabel robot actions in hindsight~\cite{laskey2017comparing}. An alternative is to cede control of the system to a human supervisor for short corrective trajectories (termed \textit{interventions}) in states where the robot policy needs assistance. 
Interventional data collection can either be human-gated~\cite{Kelly2018HGDAggerII, luo2021robust}, %
where the human monitors the policy and decides when to provide interventions, or robot-gated~\cite{hoque2021lazydagger, hoque2021thriftydagger, ifl}, %
where the robot decides when the human should provide interventions.
However, these approaches require collecting a sufficient number of human interventions for the robot to learn robust recovery.
In this work, we develop a novel data generation mechanism based on replay-based imitation \cite{mimicgen, Wen2022YouOD, Johns2021CoarsetoFineIL} in order to alleviate this burden.

\textbf{Policy Adaptation under Domain Shift.}
There are other approaches besides interactive IL for increasing policy robustness.
These include injecting noise during demonstration collection~\cite{laskey2017dart}, having human operators intentionally introduce mistakes and corrections during data collection~\cite{brandfonbrener2023visual}, and enabling policies to deal with partial observability~\cite{nguyen2022leveraging, choudhury2017learning}.
Other approaches include employing a planner to return to states that the agent has seen before~\cite{wong2022error, cideron2023get}, using Reinforcement Learning (RL) with learned rewards to help an agent adapt to new object distributions~\cite{haldar2023teach}, and using counterfactual data augmentation to identify irrelevant concepts and ensure agent behavior will not be affected by them~\cite{peng2023diagnosis}.
There are also approaches to make policies trained with RL more robust, such as domain randomization~\cite{tobin2017domain, peng2018sim}, using adversarial perturbations~\cite{mandlekar2017adversarially}, and training agents to recover from unsafe situations~\cite{thananjeyan2021recovery}. %

\revision{\textbf{MimicGen.} MimicGen \cite{mimicgen} is a recently proposed system for automatically generating task demonstrations via trajectory adaptation via leveraing known object poses. \algabbr employs a similar mechanism for synthesizing trajectories but has several key differences. Unlike MimicGen, \algabbr (1) generates interventional data rather than full demonstrations, (2) relaxes the assumption of precise object pose knowledge, which is critical to MimicGen's success, (3) integrates closed-loop policy execution that allows the robot to visit novel states during the data generation process, and (4) allows variation in not just object poses but also robot belief states about these object poses.}

%% file: figures_tex/example.tex
\begin{figure*}[t!]
  \begin{center}
    \includegraphics[width=0.95\textwidth]{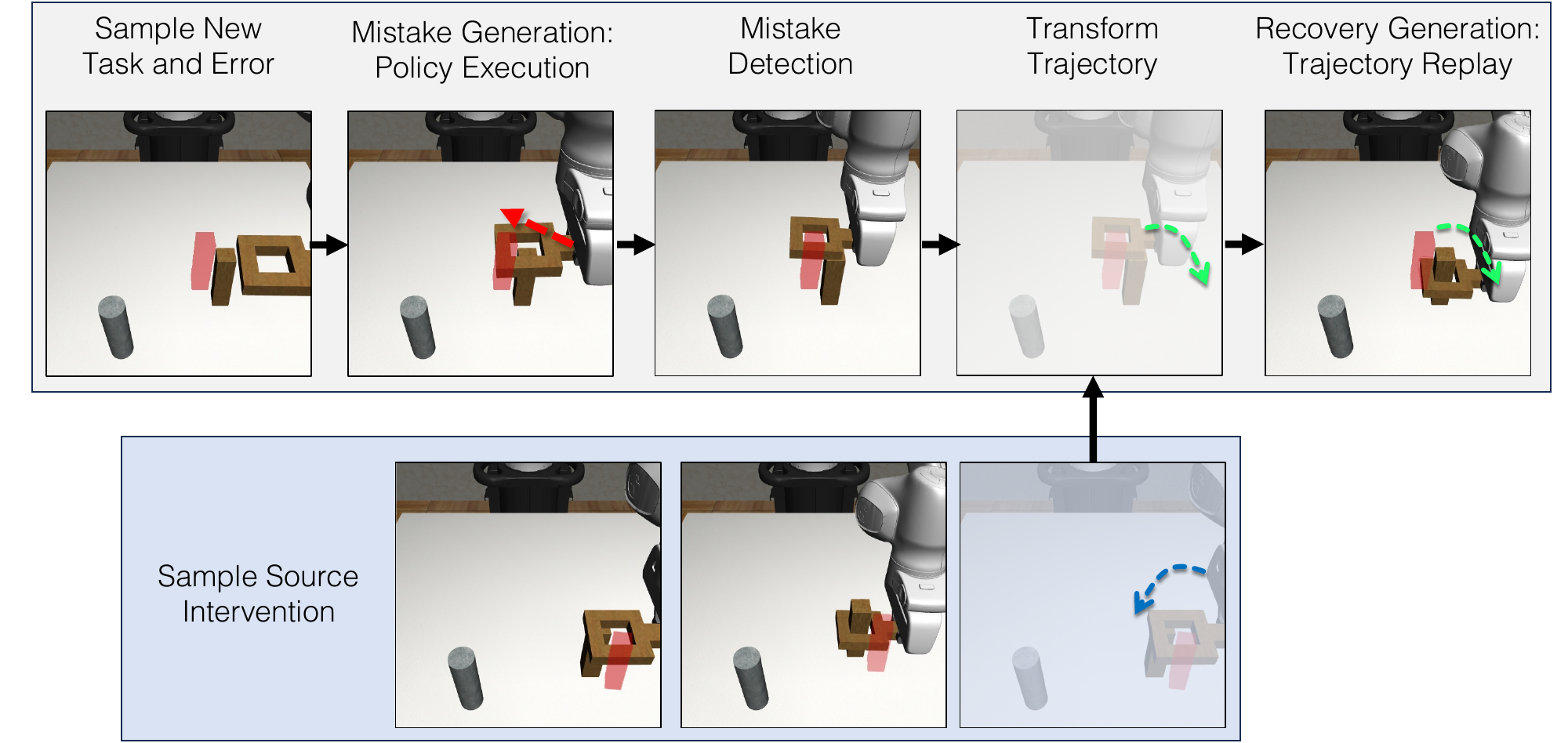}
  \end{center}
  \caption{\textbf{\algabbr Data Generation Example.} We provide an example of how \algabbr generates a new intervention. First, a new task instance is sampled with a new configuration (square peg location) and observation corruption (incorrect peg location highlighted in red). We execute the robot policy to generate mistake behavior for the new task instance. When a mistake is detected, we sample a human intervention segment from the source dataset and transform it to adapt to the current scene. Finally, we executed the transformed recovery segment in the environment.}
  \label{fig:sys_example}
  \vspace{-15pt}
\end{figure*}

%% file: chapters/ps.tex
\section{Preliminaries}\label{sec:ps}

\textbf{Problem Statement.}
We model the task environment as a Partially Observable Markov Decision Process (POMDP) with state space $S$, observation space $O$, and action space $A$. The robot does not have access to the transition dynamics or reward function but has a dataset of samples $D = \{(o,a)\}_{i=1}^N$ from an expert human policy $\pi_H: O \rightarrow A$. We assume that while the human observes observation $o$, the robot's observation is corrupted by some function $z$, yielding $z(o) = o' \in O$ (e.g., due to sensor noise or network delay).
In this work we train policies on demonstration datasets $D$ using supervised learning with the objective $\arg\min_{\theta} \mathbb{E}_{(o, a) \sim D} [-\log \pi_{\theta}(a | o)]$.

\textbf{Assumptions.}
\algabbr has assumptions similar to MimicGen~\cite{mimicgen}. 
\textbf{(Assumption 1)} the action space consists of delta-pose commands in Cartesian end effector space; 
\textbf{(Assumption 2)} the task is a known sequence of object-centric subtasks; 
\textbf{(Assumption 3)} object poses can be observed at the beginning of each subtask during data collection (but not deployment). 
\textbf{(Assumption 4)} We also assume that demonstrated recovery behavior can be explained by some component of the robot's observations $\{o'_1, o'_2, \dots\}$ during a human intervention despite corruption by $z$. Without this assumption, it would not be possible for the robot to learn a policy that maps $o'$ to $\pi_H(o)$. This information can be provided, for instance, in additional observation modalities such as force-torque sensing or tactile sensing that provide a coarse signal about an object's pose. Some settings may not require any additional information: for example, a fully closed gripper can inform the robot it must recover from a missed grasp.

\textbf{MimicGen Data Generation System.}
MimicGen~\cite{mimicgen} takes a small set of source human demonstrations $D_{src}$ and uses it to automatically generate a large dataset $D$ in a target environment. It first divides each source trajectory $\tau \in D_{src}$ into object-centric manipulation segments $\{\tau_i\}_{i=1}^M$, each of which corresponds to an object-centric subtask (Assumption 2 above). Each segment is a sequence of end effector poses. Then, to generate a demonstration in a new scene, it uses the pose of the object corresponding to the current subtask, and transforms the poses in a source human segment $\tau_i$ (with an SE(3) transform) such that the relative poses between the end effector and the object frame are preserved between the source demonstration and the new scene. It also adds an interpolation segment between the robot's current configuration and the start of the transformed segment. Then, the sequence of poses in the interpolation segment and transformed segment are executed by the robot end effector controller open-loop until the current subtask is complete, at which point the process repeats for the next subtask. We use a data generation mechanism similar to MimicGen to generate intervention trajectory segments in Section~\ref{ssec:recovery}. %

%% file: chapters/method.tex
\section{\algname}

Algorithm~\ref{alg:algorithm} displays the full pseudocode for \algname. 
It takes as input the initial state distribution $p_0$, a base dataset of demonstrations $D$, and three hyperparameters $k, m, n$. On each of one or more iterations, the system: (1) trains a policy $\pi_\theta$ on the current dataset; (2) rolls out $\pi_\theta$ for interventional data collection with human teleoperation (Sec.~\ref{ssec:inputdata}); (3) synthesizes new interventions with closed-loop policy execution and open-loop trajectory replay (Sec.~\ref{ssec:policy-exec} and Sec.~\ref{ssec:recovery}); (4) returns the new synthetic dataset.

\subsection{Interventional Data Collection}\label{ssec:inputdata}

We consider human-gated interventions \cite{Kelly2018HGDAggerII}, in which the human monitors the robot policy execution and intermittently takes control to correct policy mistakes. As in DAgger \cite{ross2011reduction}, this enables the human to demonstrate corrective recovery behavior from mistakes made by the robot policy that otherwise would not be visited in full human task demonstrations (due to distribution shift). The base robot policy $\pi_\theta$ executed during interventional data collection can come from anywhere, but is typically initialized from behavior cloning on an initial set of offline task demonstrations $D$ \cite{hoque2021thriftydagger, Mandlekar2020HumanintheLoopIL, ross2011reduction}. Each collected trajectory can be coarsely divided into robot-generated ``mistake" segments and human-generated ``recovery" segments.

\subsection{Mistake Generation: Closed-Loop Policy Execution}\label{ssec:policy-exec}

We aim to use the collected human interventions to automatically synthesize interventions for new scene configurations.
\revision{Recall that, in prior work, MimicGen generates data by executing a sequence of object-centric trajectories in an open-loop manner.}
In contrast, an appealing property of our interventional IL setting is access to the robot policy $\pi_\theta$ that is executed during interventional data collection with the human operator. 

\revision{We use this robot policy during the \textit{data generation} process to broaden the distribution of visited mistake states.}
Unlike MimicGen, 
we can execute the policy in the new scene configuration. This has two benefits: (1) rather than assuming the policy will fail in the same manner as the source trajectory, the generated mistake will reflect the genuine behavior of the policy in the new configuration, and (2) it becomes possible to generate new mistake trajectories for new corruptions of the observed object poses. For example, if sensor noise corrupts the object pose during interventional data collection, a new noise corruption can be applied during the data generation process. This allows data diversity in both object poses and the robot's erroneous beliefs about where the objects are (see Fig.~\ref{fig:sys_example}). 
\revision{The use of policy execution during data generation requires that we know when to terminate the policy execution. In our experiments, we use contact detection to determine whether or not the policy made a mistake. A more flexible option could be to use a learned classifier or robot-gated intervention criteria such as ThriftyDAgger \cite{hoque2021thriftydagger}.}

\subsection{Recovery Generation: Open-Loop Trajectory Replay}
\label{ssec:recovery}

In each episode of synthetic data generation, once we have completed policy execution and entered a new mistake state, we generate a recovery trajectory. We select a random source trajectory, segment out the human recovery portion of the trajectory, and adapt the trajectory to the current environment state. 
This adaptation consists of (1) transforming the source trajectory to the current object pose, (2) linearly interpolating in end-effector space to the beginning of the transformed trajectory, and (3) executing the transformed trajectory open-loop (see Fig.~\ref{fig:sys_example}). Note that each object-centric subtask in a single task instance can have zero, one, or multiple instances of alternating between mistake and recovery. 

\subsection{Output Filtering and Dataset Aggregation}

\revision{It is possible that the executed trajectory may not complete the task successfully. For instance, the recovery trajectory may be unable to recover from the new mistake state reached by the robot. 
Consequently, }we only keep the generated demonstration if it successfully completes the task. %
We also filter out the segment of the synthetic demonstration that corresponds to the human recovery segment; such filtering is used by common algorithms such as DAgger \cite{ross2011reduction} and HG-DAgger \cite{Kelly2018HGDAggerII} and can prevent the imitation of mistakes. 
Each filtered episode of synthetic data is aggregated into the base dataset $D$ (used to train the base policy $\pi_\theta$), and the policy is retrained on the new dataset after data generation. 
If desired, the entire process of data collection, data generation, and policy training can be iterated.

\subsection{Inter-Subtask Recovery and Offline Mode}

The \algabbr framework accommodates additional modules not considered in the main set of experiments that greatly increases its range of applications, including (1) policy recovery from more severe failure modes that revert to earlier subtasks and (2) ``offline" \algabbr, which allows humans to demonstrate mistakes intentionally~\cite{brandfonbrener2023visual}. We include experiments for these modules on the supplemental \href{https://sites.google.com/view/intervengen2024}{website}.

\begin{algorithm}[!t]
  \begin{small} %
  \caption{\algname}
  \label{alg:algorithm}
  \begin{algorithmic}[1] %
    \Declare Initial state distribution $p_0$, base dataset $D$
    \Declare Number of iterations $k$, human intervention episodes $m$, and synthesized trajectories $n$
    \Procedure{\algabbr}{$p_0, D; k, m, n$}
        \For{$i \in [1, ..., k]$} \Comment{One or more iterations}
            \State $\pi_\theta \gets \proc{train-policy}(D)$
            \State $\mathcal{D} = \emptyset $
            \For{$j \in [1, ..., m]$} \Comment{\textbf{Data Collection}}
                \State $s_0 \sim p_0$ \Comment{Sample initial state}
                \State $\tau \gets \proc{execute-policy}(s_0, \pi_\theta)$
                \State $\proc{intervene}(\tau)$ \Comment{Human intervention}
                \State $\mathcal{D} \gets \mathcal{D} \cup \tau$
            \EndFor
            \For{$j \in [1, ..., n]$} \Comment{\textbf{Data Generation}}
                \State $s_0 \sim p_0$ 
                \State $\xi \gets \proc{execute-policy}(s_0, \pi_\theta)$ 
                \State $t \gets \proc{terminate-policy}(\xi)$
                \State $\tau \sim \mathcal{D}$ \Comment{Sample source demonstration}
                \State $\tau \gets \tau[\text{human}]$ \Comment{Filter intervention}
                \State $\tau' \gets \proc{adapt}(\xi,\tau)$ \Comment{Transform trajectory} %
                \State $\xi \gets \xi \oplus \proc{replay}(\tau')$
                \If{$\proc{satisfies-goal}(\xi[-1])$}
                    \State $D \gets D \cup \xi[t:]$ \Comment{Filter intervention}
                \EndIf
            \EndFor
        \EndFor
        \State \Return $D$ %
    \EndProcedure
\end{algorithmic}
\end{small}%
\end{algorithm}

%% file: chapters/setup.tex
\section{Experiment Setup}
\label{sec:setup}

\input{figures_tex/tasks}

We consider \revision{4} tasks in the MuJoCo~\cite{mujoco} robosuite simulation environment~\cite{robosuite2020} (Fig.~\ref{fig:tasks}) \revision{and 1 physical experiment. Each task involves contact-rich manipulation via continuous control. The tasks vary in object geometry, object pose, observation error, and number of manipulation stages.} 

\textbf{Nut Insertion:} The robot must place a square nut (held in-hand) onto a square peg. The peg position is sampled in a 10\,cm x 10\,cm region at the start of each episode.

\textbf{2-Piece Assembly:} The robot must place an object into a square receptacle with a narrow affordance region. The receptacle position is sampled in a 10\,cm x 10\,cm region at the start of each episode.

\textbf{Coffee:} The robot must place and release a coffee pod into a coffee machine pod holder with a narrow affordance region. The coffee machine position is sampled in a 10\,cm x 10\,cm region at the start of each episode.

\textbf{Nut-and-Peg Assembly}~\cite{robosuite2020, Mandlekar2021WhatMI}: A multi-stage task consisting of (1) grasping a nut with a varying initial position and orientation and (2) placing it on a peg in a fixed target location. The nut is placed in a 0.5\,cm x 11.5\,cm region with a random top-down rotation at the start of each episode.

\revision{\textbf{Physical Block Grasp:} A Franka robot arm must reach a block and grasp it. The initial block position is sampled in a 20\,cm x 30\,cm region at the start of each episode.
} 
\ 
\\

\noindent\textbf{Sources of Observation Error.} In most environments, the source of observation error is \textit{sensor noise}: at test time, uniform random noise is applied to the observed position of the peg ($\pm 4$\,cm in each dimension, with at least $2$\,cm in one dimension), receptacle ($\pm 4$\,cm in each dimension, with at least $1$\,cm in one dimension), coffee machine (radial noise between $2$\,cm and $4$\,cm), \revision{and block ($\pm 1$\,cm in $x$ and $\pm 7$\,cm in $y$, with at least $2.5$\,cm in $y$)} respectively. 
In the Nut-and-Peg Assembly environment, the source of observation error is \textit{object geometry}: for an identical observed nut pose, the nut handle may exist on either of two sides of the nut. This setting corresponds to object model misspecification during pose registration.

\subsection{Experimental Setup}

\textbf{Data Collection.} For interventional data collection, we use the remote teleoperation system proposed by Mandlekar et al.~\cite{Mandlekar2020HumanintheLoopIL}.
The observation space consists of robot proprioception (6DOF end effector pose and gripper finger width) and object poses, while the action space consists of 6DOF pose deltas and a binary gripper open/close command \revision{(except for Block Grasp, which uses 3DOF position control with fixed rotation)}. For the base policy $\pi_\theta$ used in each task, we (1) collect 10 full human task demonstrations in each environment \textit{without} observation corruption (i.e., ground truth poses), (2) synthesize 1000 demonstrations with MimicGen~\cite{mimicgen}, and (3) train an off-the-shelf BC-RNN policy with default hyperparameters using the robomimic framework~\cite{Mandlekar2021WhatMI}, with the exception of an increased learning rate of 0.001~\cite{mimicgen}. 

\textbf{Data Generation.} We then deploy $\pi_\theta$ in the test environment \textit{with} observation corruption (i.e., object pose error) and collect 10 human-gated interventions. These interventions are expanded to 1000 synthetic interventions with \algabbr and aggregated with the 1000 demonstrations used to train the base policy. Finally, we train a new BC-RNN policy 
on the aggregated dataset. We report policy performance as the success rate over 50 trials for the highest performing checkpoint during training (where training takes 2000 epochs with evaluation every 50 epochs), as in~\cite{Mandlekar2021WhatMI, mimicgen}.

\textbf{Observability.} In order for demonstrated recovery behavior to be learnable (Section~\ref{sec:ps}), \algabbr and all baselines can access additional observation information in Nut Insertion, Two-Piece Assembly, Coffee, \revision{and Block Grasp} upon contact between (1) the nut and peg, (2) object and receptacle, (3) pod and pod holder, \revision{and (4) gripper and cube,} respectively. We study both the idealized case of full observability (i.e., ground truth pose) upon contact in Section~\ref{ssec:results} and partially improved observability (e.g., position of contact) in Section~\ref{ssec:analysis}. These are intended to be surrogates for sensor modalities such as force-torque sensing that can help inform the robot about the object pose when its belief is wrong. For Nut-and-Peg Assembly, we do not add additional information, as a closed gripper state is sufficient for the policy to map a missed grasp to learned recovery.

\revision{\textbf{Physical Experiment Setup.} 
We wish to evaluate whether or not policies trained on simulation data from \algabbr can retain their robustness to erroneous state estimation when they are deployed directly in the real world. 
To do this, we train a policy for the Block Grasp task in simulation and deploy it zero-shot on a physical robot. We use a Franka Research 3 robot arm and gripper and a red cube with a side length of 5\,cm. 
We use an Intel RealSense D415 depth camera and Iterative Closest Point (ICP) for cube pose estimation. %
The deployed policies output continuous control delta-pose actions at 20 Hz and do not require any real-world data or fine-tuning. 
See Figure~\ref{fig:sim2real} for images of the transfer process.}

\subsection{Baselines}

We implement and evaluate the following baselines. Each baseline corresponds to a \textit{different dataset} used to train the agent (all agents are trained with BC-RNN~\cite{Mandlekar2021WhatMI}): 

\textbf{Base:} Deploy the base policy in the test environment without any additional data or fine-tuning.

\textbf{Source Interventions} (Source Int): Deploy the base policy $\pi_\theta$, collect 10 human interventions when the policy makes mistakes, and add them to the base dataset.

\textbf{Weighted Source Interventions} (Weighted Src Int)~\cite{Mandlekar2020HumanintheLoopIL}: Same as Source Interventions, but weight the intervention data higher so that it is sampled as frequently as the base data despite its smaller quantity.

\textbf{Source Demonstrations} (Source Demo): Collect 10 full human task demonstrations in the test environment.

\textbf{MimicGen Demonstrations} (MG Demo)~\cite{mimicgen}: Same as Source Demonstrations, but use (regular) MimicGen to generate 1000 synthetic demonstrations from the initial 10.

\textbf{Policy Execution Ablation} (\algabbr\ - Policy): Augment the 10 source interventions to 1000 \algabbr interventions, but do not use policy execution to generate new mistake states. 

\input{figures_tex/sim2real.tex}

%% file: figures_tex/tasks.tex
\begin{figure*}[ht!]
  \begin{center}
    \includegraphics[width=1.0\textwidth]{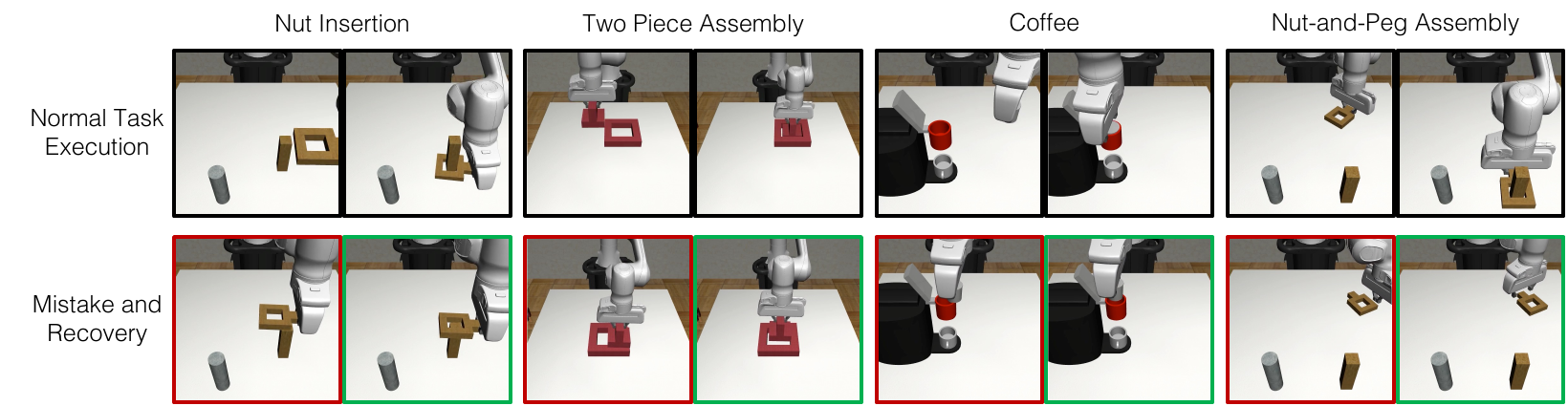}
  \end{center}
  \caption{\textbf{Tasks.} We evaluate \algabbr in several contact-rich, high-precision tasks. The top row shows normal task execution while the bottom row shows typical mistakes encountered by the agent when using inaccurate object poses (or object geometry for Nut-and-Peg Assembly) and associated recovery behaviors.}
  \label{fig:tasks}
  \vspace{-15pt}
\end{figure*}

%% file: figures_tex/sim2real.tex
\begin{figure}[ht!]
  \begin{center}
    \includegraphics[width=\linewidth]{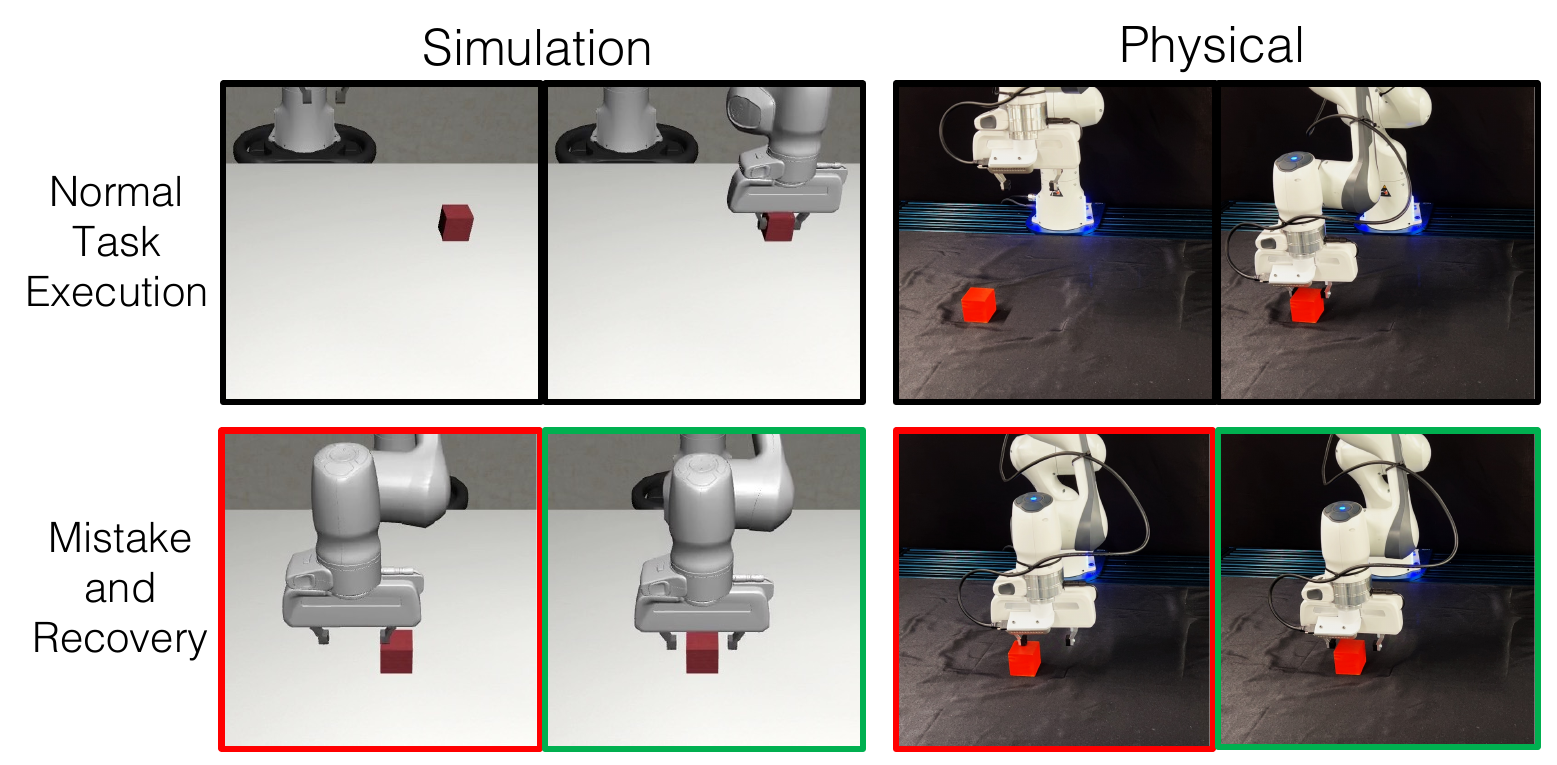}
  \end{center}
  \caption{\revision{\textbf{Sim-to-Real.} We evaluate sim-to-real transfer for a block grasping task with a Franka Panda robot. Similar to Figure~\ref{fig:tasks} we show normal task execution, typical mistakes due to inaccurate object poses, and associated recovery for the simulation and real world environments. The results show that \algabbr can facilitate sim-to-real transfer of learned control policies, and that these policies retain robustness to erroneous perception.}} %
  \label{fig:sim2real}
  \vspace{-5pt}
\end{figure}

%% file: chapters/experiments.tex
\section{Experiments}
\label{sec:exps}

\label{ssec:results}
In this section, we summarize the key takeaways from the comparisons presented in Tables~\ref{tab:main} and ~\ref{tab:nuthandle}.

\textbf{\algabbr vastly improves policy robustness under pose estimation error.} In Table~\ref{tab:main}, we observe that \algabbr improves policy performance by 3.5$\times$, 10.7$\times$, and 39$\times$ over the base policy in Nut Insertion, 2-Piece Assembly, and Coffee respectively, despite only collecting 10 human interventions.

\textbf{\algabbr significantly improves upon naïve uses of an equivalent amount of full human demonstration data.} \algabbr consistently outperforms human demonstrations collected at test time (Source Demo, Table~\ref{tab:main}) by 56\%-68\%. Even if these demonstrations are expanded by $100\times$ with MimicGen (MG Demo), \algabbr still outperforms by 34\%-62\%. Since the human's observability does not match the robot's, the human can teleoperate toward the true object poses. Thus, the robot does not observe any recovery behavior in the offline data.

\textbf{\algabbr significantly improves upon naïve uses of an equivalent amount of interventional human data.} Source Int in Table~\ref{tab:main} underperforms \algabbr by 58\%-70\%. While helpful, with only 10 human interventions, the data is insufficient to learn robust recovery under pose error. This remains the case even if the intervention data is weighted higher, in which case the agent overfits to the 10 interventions and underperforms \algabbr by 48\%-74\%. With the same budget of interventional human data, \algabbr can generate much richer coverage of the distribution of mistakes under the base policy.

\textbf{\algabbr significantly improves upon naïve uses of MimicGen.} We observe a significant 34\%-62\% improvement over MimicGen on full task demonstrations (MG Demo, Table~\ref{tab:main}). We also observe that the policy execution component (Section~\ref{ssec:policy-exec}) boosts performance by 12\%-38\% respectively over the ablation, indicating that expanding the mistake distribution is valuable. While the ablation dataset covers variation in the object pose, it does not cover variation in the error; only the 10 mistake segments in the source dataset are available. This shows that the novel components we introduced in \algabbr are crucial for high performance.

\textbf{\algabbr is useful across different environments.} While 2-Piece Assembly and Coffee have narrower tolerance regions than Nut Insertion that lower success rates across the board (16\%-20\% for the base policy, 30\%-48\% for other baselines, and 18\%-28\% for \algabbr), the relative performance of \algabbr
remains consistent across environments: \algabbr outperforms all baselines by 12\%-76\% in Nut Insertion, 18\%-64\% in 2-Pc Assembly, and 38\%-78\% in Coffee.

\textbf{\algabbr is useful across different sources of observation error.} Results for the Nut-and-Peg Assembly task with object geometry error are in Table~\ref{tab:nuthandle}. We evaluate each policy with 50 evaluations of each of the two possible geometries. Base and Source Int attain perfect performance on the original geometry but struggle with the alternate geometry (0\%-6\% performance). MG Demo has the opposite issue: since it consists of test-time demonstrations with the alternate geometry, it can attain perfect performance on the alternate but 0\% on the original. A mixture of full demonstrations on both geometries (Base + MG Demo) attains an even $60\%$ and $64\%$; since it does not observe recovery behavior it must guess between the two object geometries and has difficulty performing much higher than the 50\% expected value of random chance. Finally, \algabbr maintains $92\%$ performance on the original geometry but also learns to recover when missing its grasp due to the alternate geometry (88\%), leading to a 28\%-40\% improvement in the average case over baselines. See the \href{https://sites.google.com/view/intervengen2024}{website} for videos.

\revision{\textbf{\algabbr facilitates sim-to-real transfer of learned control policies, and these policies retain robustness to erroneous state estimation.} In Table~\ref{tab:sim2real} we observe that state-based policies for the Block Grasp task deployed zero-shot on the physical system perform similarly to simulation. By improving robustness to incorrect pose estimation, \algabbr facilitates sim-to-real transfer for state-based policies, which are easier to transfer across visual domain gaps than image-based policies but rely on accurate perception. \algabbr outperforms baselines by 14\%-94\% in simulation and 30\%-90\% in real world trials, suggesting learned recovery behaviors can transfer to real. The policy is also robust to physical perturbations, dynamic object pose changes, and visual distractors; see the \href{https://sites.google.com/view/intervengen2024}{website} for videos.}

\subsection{Analysis}\label{ssec:analysis}

In this section, we present further analysis on various aspects of \algabbr.

\textbf{How is agent performance affected as observability decreases?} 
For Nut Insertion, we replace true pose information upon contact with the mean position of the first contact between the nut and peg; for 2-Piece Assembly, we provide the unit vector in the direction of the true pose at the first point of contact. Table~\ref{tab:partial} in comparison with Table~\ref{tab:main} shows that, as expected, a degradation in observability results in a degradation in agent performance. However, \algabbr performance falls by only 4\%-8\%, indicating partial observability can be sufficient to ground recovery behavior. An important direction for future work is investigating raw real-world perception signals such as force-torque sensing.

\input{figures_tex/table_core}
\input{figures_tex/table_nut_handle}
\input{figures_tex/table_partial}
\input{figures_tex/table_real}

\textbf{How does performance vary across training seeds?}
\algabbr in the (full observability) Nut Assembly task attains 98\%, 100\%, and 98\% for 3 training seeds, indicating stability across runs (more evidence on supplemental \href{https://sites.google.com/view/intervengen2024}{website}).

\textbf{How does synthetic \algname data compare to an equal amount of human data?} 
In 2-Piece Assembly, 100 \algabbr interventions (from 10 human interventions) attain 24\% while 100 human interventions attain 46\%. Both improve upon 10 human interventions, which only attains 6\% (Table~\ref{tab:main}). 
However, 1000 \algabbr interventions from 10 human interventions (70\%) can outperform 100 human interventions, and 100 human interventions take significantly more human time and effort to collect than 10 human interventions (29.9 minutes instead of 3.6 minutes).

\textbf{How does performance scale with the amount of synthetically generated interventions?}
With the same 10 human source interventions in 2-Piece Assembly, an agent trained on 200 synthetic \algabbr interventions attains 34\%, 1000 interventions attains 70\% (Table~\ref{tab:main}), and 5000 interventions attains 88\%. This suggests performance scales with dataset size, at the cost of additional data generation time.

%% file: figures_tex/table_core.tex
\begin{table}[t!]
    \centering
    \begin{tabular}{l | r r r}
        \textbf{Dataset} & \textbf{Nut Insertion} & \textbf{2-Pc Assembly} & \textbf{Coffee}
        \\ \hline 
        Base & 22\% & 6\% & 2\% \\
        Source Int & 40\% & 6\% & 10\% \\
        Weighted Src Int~\cite{Mandlekar2020HumanintheLoopIL} & 50\% & 16\% & 6\% \\
        Source Demo & 42\% & 12\% & 12\% \\
        MG Demo~\cite{mimicgen} & 64\% & 16\% & 18\% \\
        \algabbr \ - Policy (Ours) & 86\% & 52\% & 42\% \\
        \algabbr (Ours) & \textbf{98\%} & \textbf{70\%} & \textbf{80\%} \\
    \end{tabular}
    \caption{Results in three simulation domains with noisy pose estimation and full observability upon contact. \algabbr outperforms baselines across environments.}
    \label{tab:main}
    \vspace{-5pt}
\end{table}

%% file: figures_tex/table_nut_handle.tex
\begin{table}[t!]
    \centering
    \begin{tabular}{l | r r r}
        \textbf{Dataset} & \textbf{Geometry 1} & \textbf{Geometry 2} & \textbf{Mixture} \\ \hline 
        Base & \textbf{100\%} & 0\% & 50\%  \\
        Source Int & \textbf{100\%} & 6\% & 53\% \\
        MG Demo~\cite{mimicgen} & 0\% & \textbf{100\%} & 50\% \\
        Base + MG Demo & 64\% & 60\% & 62\%   \\
        \algabbr & 92\% & 88\% & \textbf{90\%} \\
    \end{tabular}
    \caption{Results in the Nut-and-Peg Assembly experiment. While baselines typically overfit to one geometry or struggle with disambiguating the two, \algabbr attains high performance on the mixture of geometries.}
    \label{tab:nuthandle}
    \vspace{-5pt}
\end{table}

%% file: figures_tex/table_partial.tex
\begin{table}[t!]
    \centering
    \begin{tabular}{l | r r}
        \textbf{Dataset} & \textbf{Nut Insertion} & \textbf{2-Pc Assembly} \\ \hline
        Base & 26\% & 6\%  \\
        Source Int & 40\% & 6\%  \\
        MG Demo~\cite{mimicgen} & 46\% & 22\%  \\
        \algabbr \ - Policy & 68\% & 42\% \\
        \algabbr & \textbf{90\%} & \textbf{66\%} \\
    \end{tabular}
    \caption{Additional evaluation in two domains with partially improved (rather than full) observability upon contact.}
    \label{tab:partial}
    \vspace{-5pt}
\end{table}

%% file: figures_tex/table_real.tex
\begin{table}[t!]
    \centering
    \begin{tabular}{l | r r}
        \textbf{Dataset} & \textbf{Simulation} & \textbf{Real} \\ \hline
        Base & 6\% & 0\%  \\
        Source Int & 26\% & 10\%  \\
        MG Demo~\cite{mimicgen} & 42\% & 50\%  \\
        \algabbr \ - Policy & 86\% & 60\% \\
        \algabbr & \textbf{100\%} & \textbf{90\%} \\
    \end{tabular}
    \caption{\revision{Sim-to-real results for the block grasping task in simulation (50 trials) and zero-shot evaluation of these policies in the real world (10 trials).}}
    \label{tab:sim2real}
    \vspace{-10pt}
\end{table}

%% file: chapters/conclusion.tex
\section{\revision{Conclusion}}

We present \algname (\algabbr), a data generation system for corrective interventions that cover a large distribution of policy mistakes given a small number of source human interventions. %
We show that training on synthetic data generated by \algabbr compares favorably to collecting more human demonstrations and interventions in terms of both policy performance and human effort. 

\revision{
Although \algabbr improves on MimicGen and reduces its reliance on accurate pose estimation, \algabbr shares some of its limitations. Specifically, we consider only quasi-static tasks with rigid body objects, and we assume valid interventions can be synthesized by transforming source trajectory data. 

Future work involves applying \algabbr in settings with force-torque sensing to improve behavioral adaptation for contact-rich and high-precision tasks.
\algabbr can also be used to rapidly adapt policy behavior toward individual human preferences over how a manipulation task should be carried out without extensive data collection. 
Finally, \algabbr can also be applied to facilitating sim-to-real transfer of IL policies by acting as a domain randomization~\cite{tobin2017domain} procedure. 
Namely, while RL algorithms can autonomously learn adaptations to dynamical domain randomization, IL typically requires generating new human behavior for these variations. \algabbr may dramatically reduce the data requirements and enable policies to deal with such variations with only a handful of corrective behaviors.
}

%% file: acks.tex
\section*{Acknowledgements}

This work was made possible with the support of the NVIDIA Seattle Robotics Lab. We especially thank Ankur Handa for valuable discussions, Ravinder Singh for IT help, and Sandeep Desai for robot hardware support.